# APPLICATION OF FUZZY SYSTEM IN SEGMENTATION OF MRI BRAIN TUMOR

MRIGANK RAJYA

CSE, LINGAYA'S UNIVERSITY

LIMAT, FARIDABAD INDIA

NEW DELHI, INDIA

mrigankrajya@gmail.com

SONAL REWRI

CSE, LINGAYA'S UNIVERSITY

LIMAT, FARIDABAD INDIA

NEW DELHI, INDIA

pretty_sonal16@yahoo.co.in

SWATI SHEORAN

CSE, LINGAYA'S UNIVERSITY

LIMAT, FARIDABAD INDIA

NEW DELHI, INDIA

swati.sheoran @gmail.com

**ABSTRACT**

**Segmentation of images holds an important position in the area of image processing. It becomes more important while typically dealing with medical images where presurgery and post surgery decisions are required for the purpose of initiating and speeding up the recovery process. Segmentation of 3-D tumor structures from magnetic resonance images (MRI) is a very challenging problem due to the variability of tumor geometry and intensity patterns. Level set evolution combining global smoothness with the flexibility of topology changes offers significant advantages over the conventional statistical classification followed by mathematical morphology. Level set evolution with constant propagation needs to be initialized either completely inside or outside the tumor and can leak through weak or missing boundary parts. Replacing the constant propagation term by a statistical force overcomes these limitations and results in a convergence to a stable solution. Using MR images presenting tumors, probabilities for background and tumor regions are calculated from a pre- and post-contrast difference image and mixture modeling fit of the histogram. The whole image is used for initialization of the level set evolution to segment the tumor boundaries.**

## 1. INTRODUCTION

Segmentation is an important technique used in image processing to identify objects in an image. We are interested in segmentation techniques that can be applied in a robust and efficient way to both image and mesh data. Mesh data is frequently unstructured, this precludes the direct application of techniques that are originally developed for the more structured image data. One solution to this problem is the use of techniques such as level sets or implicit active contours. The idea behind active contours, or deformable models, for image segmentation is quite simple. The user specifies an initial guess for the contour, which is then moved by image driven forces to the boundaries of the desired objects. In such models, two types of forces are considered – the internal forces, defined within the curve, are designed keep the model smooth during the deformation process, while the external forces, which are computed from the underlying image data, are defined to move the model toward an object boundary or other desired features within the image [1-2]. There are two forms of deformable models. In the parametric form, also referred to as snakes, an explicit parametric representation of the curve is used. This form is not only compact, but is robust to both image noise and boundary gaps as it constrains the extracted boundaries to be smooth. However, it can severely restrict the degree of topological adaptability of the model, especially if the deformation involves splitting or merging of parts. In contrast, the implicit deformable models, also called implicit active contours or level sets, are designed to handle topological changes naturally. However, unlike the parametric form, they are not robust to boundary gaps and suffer from several other deficiencies as well [3-4]. Among the various image segmentation techniques, the level set method offer a powerful approach for image segmentation since it can handle any of the cavities, concavities, splitting/merging and convolution. It has been used in wide fields including medical image processing [4-6]. However, despite all of the advantages, which this method can provide, it requires the prior choice of the most important parameters such as the initial location of seed point, the appropriate propagation speed function and the degree of smoothness. The traditional methods only depend on the contrast of the points located near the object boundaries, which cannot be used for the accurate segmentation of complex medical images [5-6]. Segmentation of volumetric image data is still a challenging problem and solutions are based on either simple intensity thresholding or model-based deformation of templates. The former implies that structures are well separated by unique intensity patterns, whereas the latter requires model templates characteristic for the shape class. Snakes [7] are appealing to users as they require only a coarse initialization and converge to a stable, reproducible boundary. Snakes based on level set evolution are especially appealing for volumetric data processing [8-10]. The formalism can be naturally extended from 2D to higher dimensions and the result zero level set offers flexible topology. Level set with fixed propagation direction is either initialized inside or outside sought objects and the propagation force is opposed by a strong gradient magnitude at image discontinuities. At locations of missing or fuzzy boundaries, the internal force is often strong enough to counteract global smoothness and leaks through these gaps. Thus, there is no convergence and the evolution has to be halted manually. This observation led to a new concept of region competition, where two adjacent regions compete for the common boundary [11-12] and
are additionally constrained by a smoothness term. The driving problem discussed in this paper is the segmentation of 3-D brain tumors from magnetic resonance image data. Tumors vary in shape, size, location and internal texture and tumor segmentation is herefore known to be a very challenging and difficult problem. Intensity thresholding followed by erosion, connectivity and dilation is a common procedure but only applicable to a small class of tumors presenting simple shape and homogeneous interior structure. Warfield et al. [13-15] suggested a methodology based on elastic atlas warping for brain extraction and statistical pattern recognition for brain interior structures. The intensity feature was augmented by a distance from the boundary feature to account for overlapping probability density functions. This method was found to be successful for simple-shaped tumors with homogeneous texture.

## 2. PROPOSED SEGMENTATION METHODOLOGY

The method starts with an intensity-based fuzzy classification of voxels into tumor and background classes. The details of this initial classification are covered in Section III-A. The tumor probability map is used to locally guide the propagation direction and speed of a level-set snake. The tumor probability map is also used to derive an automatic initialization of the snake. Image forces are balanced with global smoothness constraints to converge stably to a smooth tumor segmentation of arbitrary topology. In this work we use T2-weighted pre- and post-contrast 3D images.

### A. 3D FILTERING

Preprocessing is an important task since the segmentation will be better if there is minimal image noise. For this purpose two filters are used, an anisotropic filter proposed by Perona and Malik [16] and anisotropic filter based on min/max flow scheme proposed by Malladi and Sethian [17]. The second filter is a scheme for image enhancement and noise removal based on level set theory. This filter was modified and extended to third dimension in order to eliminate the speckle noise and to keep the most significant edges of the volumes [18].

### B. ACTIVE CONTOUR (SNAKES)

This algorithm, first introduced by Kass et al., deforms a contour to lock onto features of interest within an image [19]. Usually the features are lines, edges and/or object boundaries. Kass et al. named their algorithm snakes, because the deformable contours resemble snakes as they move. Given an approximation of the boundary of an object in an image, an active contour model can be used to find the actual boundary. Active contour models should be able to find the intracranial boundary in MR images of the head when an initial guess is provided by a user or by some other method, possibly an automated one. Active contour models offer a solution to a variety of tasks in image analysis and machine vision. Active contour models can be used in image segmentation and understanding and are also suitable for analysis of dynamic image data or 3D image data. The active contour model or snake is defined as energy – minimizing spline- the snakes energy depends on its shape and location within the image. Local minima of this energy then correspond to desired image properties. Snakes may be understood as a special case of a more general technique of matching a deformable model to an image by means of energy minimization. Snakes do not completely solve the problem of finding contours in images since they depend on other mechanism such as user interaction or the output from higher-level information processing. This interaction must specify an approximate shape and starting position for the snake somewhere near the desired contour. A priori information is then used to push the snake toward an appropriate solution figure1 and figure 2. Unlike most other image models, the snake is active, always minimizing its energy function, therefore exhibiting dynamic behavior [20].

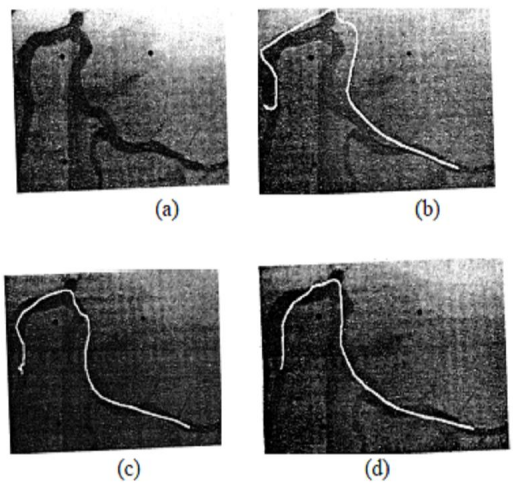

Figure 2: Snake based detection in an angiographic x-ray image of a pig heart. (a) Original angiogram (b) initial position of the snake (c) snake deformation after 4 iterations (d) final position of the snake after 10 iterations.

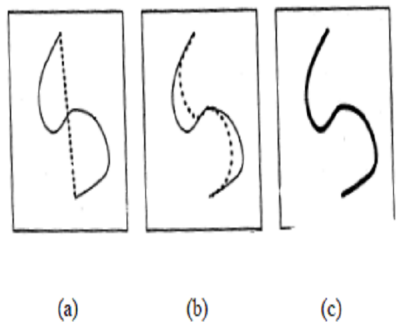

Figure 1: Active contour model: (a) Initial snake position dotted line, (b) & (c) are the steps of snake's for energy minimization.

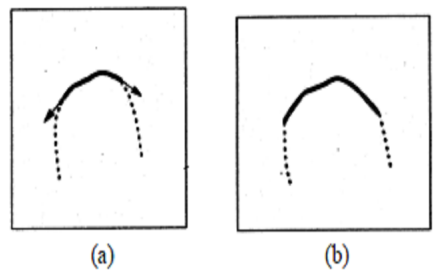

Figure 3: Snake growing (a) Lengthening in tangent direction (b) energy minimization after a growing step.

Normally, there are two types of snake models: the implicit ones and the parametric ones. Implicit models consist basically of embedding the snake as the zero level set of a higher dimensional function and solving the corresponding equation of motion. Such methodologies are best suited for the recovery of objects with complex shapes and unknown topologies. However, due to the higher dimensional formulation, implicit models are not as convenient as the parametric ones, for shape analysis and visualization and for user interaction. The parametric snake models consist basically of an elastic curve or surface which can dynamically conform to object shapes in response to internal forces and external forces. These forces can be the result of a functional global minimization process or based on local information. Such an approach is more intuitive than the implicit models. Its mathematical formulation makes it easier to integrate image data, an initial estimate, desired contour properties and knowledge-based constraints, in a single extraction process. However, parametric models also have their limitations. First, most of these methods can only handle topologically simple objects. The topology of the structures of interest must be known in advance since the mathematical model can not deal with topological changes without adding extra machinery. Second, parametric snakes are too sensitive to their initial conditions due to the no convexity of the energy function and the contraction force which arises from the internal energy term. Several works have attempted to address these limitations. By introducing the idea of snake growing Euler Lagrange equation numerical instability is eliminated. This method starts with a single snake which then divides into pieces. After each growing step, the energy of each snake is minimized where ends are pulled to the true contour as shown in figure 3. There exits a variety of other energy minimizing approaches too [20]. The Kass paper is the foundation for all active contour models and outlines the underlining equation of all snakes. In this study, we implemented a geodesic deformable model provided by R. Kimmel, V. Caselles et al. [21] for object segmentation which allows connecting classical snakes based on energy minimization and geometric active contours based on the theory of curve evolution.

### C. IMAGE FORCES AND SMOOTHNESS CONSTRAINTS

In a deformation model scheme, the model is driven by image forces and constrained by prior information on the shape of the model. The image forces are generally governed by the gradient magnitude and the shape prior which is a form of smoothness. In the level-set snake's framework, the image forces and smoothness constraints are simply separate terms in the partial differential equation governing the evolution of the implicit function defining the snake. The traditional balloon snakes, with a constant propagation term, have the issue that the propagation term has a fixed sign meaning that the balloon can only grow or shrink. Hence the snake must be initialized either completely inside the target object, or completely circumscribing the target object. In our snake model, the propagation term is locally modulated by a signed image force factor between -1 and +1, causing the snake to shrink parts outside the tumor and expand parts inside the tumor. Hence our snake can be initialized partially inside and partially outside the tumor and it is more robust to initialization. The snake guided by image forces would leak into many small noisy structures in the image that are not part of the tumor. To avoid this, the standard way to constrain level-set snakes is to apply mean curvature flow to the snake contour. In the level-set formalism this is easily done by adding a term to the snake evolution equation. A smoothing is used to the implicit function in order to aid numerical stability of the algorithm.

### D. LEVEL SET SNAKE METHOD

A physical shaped model is introduced by Malladi et al. [22]. They have developed a propagation model based on a non-intersecting closed curve where the propagation speed depends on the curvature. Here also, the level set splits the space into two regions (inside and outside), it is called "interface" in Sethian literature [23]. The front propagates itself inside the image adapting itself to the wall of the structure or 3D object. This technique solves two of the problems of snakes [24]: firstly, it allows the segmentation of objects with many bifurcations and protuberances and recovers complex shapes inside the image. Secondly, it is not necessary to know a priori the topology of an object in order for it to be recovered.

### 3. IMPLEMENTATION OF TUMOR SEGMENTATION

### A. TUMOR PROBABILITY MAP

The probability map is a scalar field on the image which specifies, voxel by voxel, a probability that the given voxel belongs to the tumor. The histogram of the difference image (voxel-by-voxel) is shown in figure 4 which clearly shows a symmetric distribution around zero and a second distribution related to regional changes caused by the contrast agent. The second distribution is asymmetric but strictly on the positive axis, which relates to larger regions accumulating a small amount of contrast and very small regions strongly highlighted by contrast. In this work, we fit the histogram by a mixture density of two distributions, a Gaussian function to model small differences around zero and a Poisson distribution to model the changes due to contrast. We use the nonlinear fit package provided by mathematics. The scalar field derived from the posterior probability with range [-1, 1] is passed into the level-set algorithm as the probability map p (A) - p (B). For computational speed, the datasets are subsequently cropped to the region around the tumor although the algorithm works-well on the whole brain. The resulting threshold is shown as a dotted vertical line. Posterior probabilities for background (solid) and tumor (dotted) as a function of difference image value are shown in the right figure. The probability map from the difference image is rather noisy and includes many regions that should not be considered part of the tumor. It is known that blood vessels and bone marrow also take up gadolinium. In addition, the ring-shaped gadolinium enhancement common in glioblastoma tumors results in misclassification of many voxels in the center of the tumor. It is clear that intensity-based tissue classification alone is insufficient for satisfactory segmentation of the tumor. The level-set snake uses this fuzzy classification heavily for its image forces with addition smoothness constraints [26].

### B. INITIALIZATION

Many conventional snakes require the user to initialize the snake with a bubble either completely inside or completely outside the object to be segmented. The competition snake does not have any constant-velocity inward/ outward propagation force, so it can be initialized with some parts of the contour inside and some parts outside the object to be segmented. The competition snake is more robust to variable initialization. We choose the level zero set of the tumor probability map, where P (tumor) = P (non-tumor), as the initialization. The implicit function is initialized to the distance map of the initial contour. In this way we have an automatic initialization of the snake, the user is not required to define seed points completely inside the object to be segmented.

### 4. RESULTS

The specifications of the images used are: image type-T2 (axial), thickness of image slice-0.5 cm, number of images per data set –23, Image format- Gif. Figure 5 and Figure 6 shows brain MRI original images and images after final contour for 300 iterations. Figure 7 shows images after segmentation. The method requires that both enhancing and non-enhancing parts of the tumor have similar appearance characteristics in the T1 pre- contrast intensities and the T2 intensities. Figure 7 shows the implicit function and the level set snake at several stages in the segmentation of two tumor datasets. The initialization of the snake corresponds to a simple intensity windowing of the image dataset and shows the difficulty of segmenting tumors with the standard erosion-connectivity-dilation morphological operators. After 300 iterations the snake segmentation is completed. The balancing force p (A)-p (B) makes the snake very stable and it does not leak into neighboring structures. Classical snakes that use the gradient magnitude as image force often have difficulty when the boundary of the object to be segmented has gaps with weak step edges in the image and they have a tendency to leak through these gaps when pushed by a constant-sign propagation force. The user time required for setting up the segmentation parameters for any tumor size is normally less than 5 minutes. The level-set procedure was successfully run on two tumor datasets and compared with manual segmentation by an expert radiologist. The output of each segment is a binary image on the same voxel grid as the original MRI. We used the VALMET [27] image segmentation validation framework to examine various metrics of agreement of segmentation. The results for two tumor datasets are shown in Table1.

| Data set | Tumor Volume (cm³) | Manual Volume (cm³) | Agreement % |
|---|---|---|---|
| First | 50.6548 | 50.9003 | 99.51 |
| Second | 50.5146 | 53.7560 | 93.97 |

Table1: Comparison of automatic segmentation with manual segmentation.

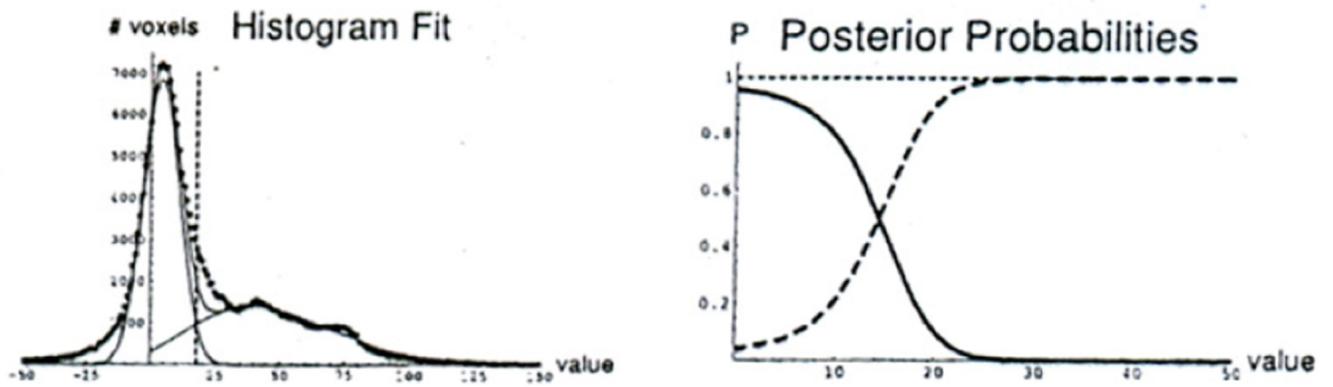

Figure 4: Histogram of post- and pre-contrast difference image fitted with two distributions (left) and posterior probabilities (right)

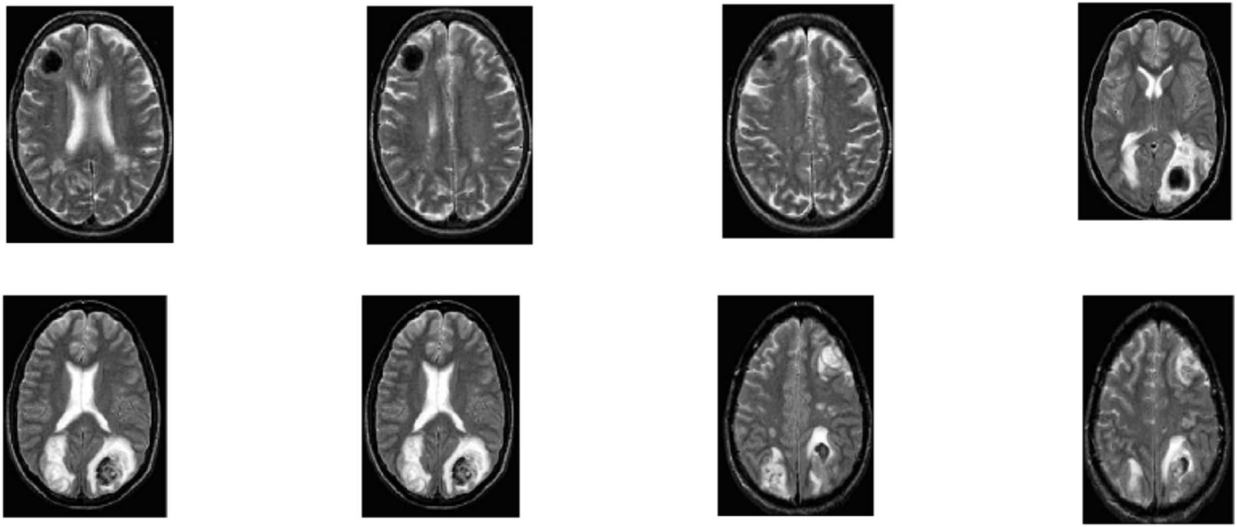

Figure 5: Brain MRI original images: 11, 12, 13, 21, 22, 23, 24, 25.

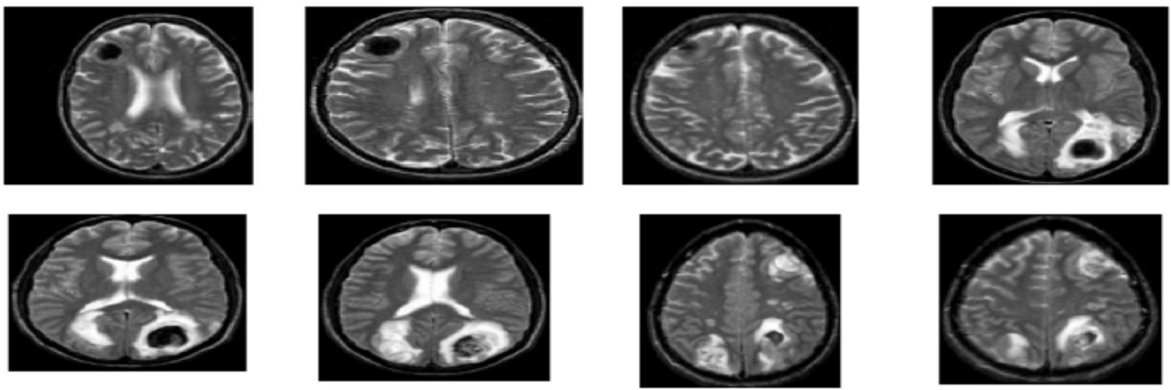

Figure 6: Images after final contour for 300 iterations: 11, 12, 13, 21, 22, 23, 24, 25.

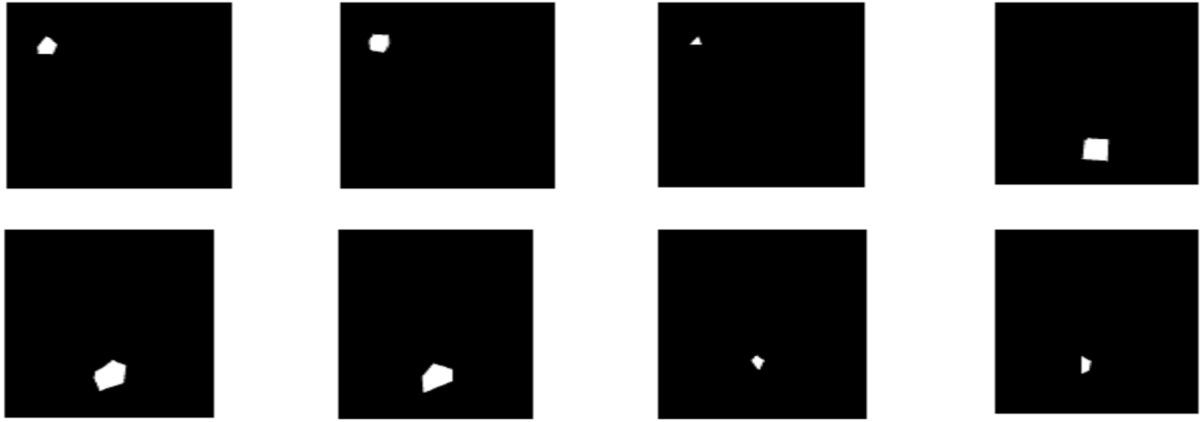

Figure 7: Images after segmentation: 11, 12, 13, 21, 22, 23, 24, 25.

## 5. CONCLUSIONS

Tumor volume is an important diagnostic indicator in treatment planning of brain tumors. The measurement of brain tumor volume could assist tumor staging for effective treatment surgical planning. Imaging plays a central role in the diagnosis and treatment planning of the brain tumor. In this study a semi-automated system for brain tumor volume measurements is developed based on MR imaging. This method is applied to 8-tumor containing MRI slices from 2 brain tumor patients' data sets and satisfactory segmentation results are achieved. We demonstrate a stable, 3D level-set evolution framework applied to automatic segmentation of brain tumors in MRI, using a probability map of tumor versus background to guide the snake propagation. A nonlinear fit of a mixture model to the histogram provides a fuzzy classification map of gadolinium-enhancing voxels and this probability map is used to guide the propagation of the snake. The snake is very stable and converges in 300 iterations without leaking. The snake is also robust to initialization. The segmentation results uses an automatic initialization at the p (A) = p (B) boundary between tumor and non-tumor regions. However, preliminary tests with various initializations indicate that snake can grow into the entire tumor even when the initialization covers only a small portion of the tumor. The automatic method has a lower level of agreement with the human experts compared to the semi automatic method. But the semiautomated method generates results that have higher level of agreement with the manual raters. Preliminary comparisons demonstrate that the semi-automatic segmentation comes close to the manual expert segmentation. Further work is in progress to investigate the sensitivity towards initialization and parameter settings on a larger set of tumor datasets.